\documentclass[letterpaper, 10 pt, conference]{ieeeconf}  

\IEEEoverridecommandlockouts                              

\usepackage{graphicx} 
\usepackage[font=small,labelfont=bf]{caption}
\usepackage{float}
\usepackage{amsmath} 
\usepackage{amsfonts,amssymb}
\usepackage[dvipsnames]{xcolor}
\usepackage{algorithm, algorithmic}
\usepackage{tabularx, booktabs, makecell}
\usepackage{tabularray}
\usepackage{siunitx}
\usepackage{pifont}
\renewcommand{\baselinestretch}{0.99}
\usepackage{cite}
\usepackage[most]{tcolorbox}
\usepackage{lipsum} 

\tcbset{
  promptstyle/.style={
    colback=gray!10,    
    colframe=gray!70,   
    fonttitle=\bfseries,
    boxrule=0.5pt,
    arc=2mm,           
    left=2mm,
    right=2mm,
    top=1mm,
    bottom=1mm,
    enhanced,
    breakable          
  }
}

\usepackage[draft]{hyperref}
\usepackage{capt-of}
\usepackage[nospread,noshrink]{cuted} 
\usepackage{etoolbox}
\usepackage{needspace}
\usepackage{CJKutf8} 
\usepackage{url}

\newcommand{\FigureSlot}[3]{%
  \begingroup
  \setlength{\fboxsep}{0pt}%
  \setlength{\fboxrule}{0.4pt}%
  \leavevmode\vbox to #2{%
    \vfil
    \hbox to #1{%
      \hfil
      \IfFileExists{#3}{%
        \includegraphics[width=#1,height=#2,keepaspectratio]{#3}%
      }{%
        \color{black!35}%
        \vbox{%
          \hrule height\fboxrule
          \hbox{%
            \vrule width\fboxrule
            \vbox to \dimexpr#2-2\fboxrule\relax{%
              \vfil
              \hbox to \dimexpr#1-2\fboxrule\relax{\hfil}%
              \vfil
            }%
            \vrule width\fboxrule
          }%
          \hrule height\fboxrule
        }%
      }%
      \hfil
    }%
    \vfil
  }%
  \endgroup
}

\makeatletter
\@ifpackagelater{cuted}{2022/01/01}{}{%
  \patchcmd{\@addviper}
    {\unvbox\hold@viper\cuted@@tempbox@var}
    {\unvbox\hold@viper\unvbox\cuted@@tempbox@var}
    {}{\PackageWarningNoLine{RoboAssist}{Please update the cuted package}}%
  \patchcmd{\@addviper}
    {\setbox\cuted@@tempbox@var\vbox}
    {\ht@viper=\cuted@@tempdim@b\relax
     \setbox\cuted@@tempbox@var\vbox}
    {}{}%
}
\newdimen\LargestInlineFigure
\newcommand{\ReserveWideFigureSpace}[1]{%
  \par
  \begingroup
    \@tempdima=\pagetotal
    \if@firstcolumn\else\advance\@tempdima by\@colht\fi
    \divide\@tempdima by 2
    \advance\@tempdima by .5\LargestInlineFigure
    \advance\@tempdima by 3\baselineskip
    \advance\@tempdima by #1\relax
    \advance\@tempdima by 2\stripsep
    \ifdim\@tempdima>\@colht\clearpage\fi
  \endgroup
}
\newcommand{\ReserveFigureSpace}[1]{%
  \par
  \begingroup
    \@tempdima=#1\relax
    \ifdim\@tempdima>\@colht
      \clearpage 
    \else
      \Needspace{#1}%
    \fi
  \endgroup
}
\newcommand{\FinishRunInHeading}{%
  \if@noskipsec\leavevmode\par\fi
  \par
}
\makeatother

\newsavebox{\SingleFigureBox}

\newsavebox{\WideFigureBox}

\makeatletter
\def\algbackskip{\hskip-\ALG@thistlm}
\makeatother

\title{\LARGE \bf
RoboAssist: Interactive Human–Humanoid Planning for Long-Horizon Surgical Assistance
}

\author{
Jingwei~Jia$^{1}$,
Keyu~Zhou$^{1}$,
Jiewei~Wang$^{1}$,
Peisen~Xu$^{2}$,
Xingyuan~Zhou$^{3}$,\\
Liang~Wang$^{2}$,
Jiming~Chen$^{2}$,
Gaofeng~Li$^{2}$,
Jin~Wang$^{4*}$,
and Shunlei~Li$^{1*}$%
\thanks{†This research was supported by Zhejiang Provincial Natural Science Foundation of China under Grant No. ZCLQN26F0306.}
\thanks{$^{1}$Hangzhou Dianzi University, China.}%
\thanks{$^{2}$Zhejiang University, China.}%
\thanks{$^{3}$New York University, USA.}%
\thanks{$^{4}$Italian Institute of Technology (IIT), Italy.}%
\thanks{$^{*}$Corresponding authors: Jin Wang
(\texttt{wangshin.v@gmail.com}) and Shunlei Li
(\texttt{shunlei.li@outlook.com}).}%
}

\IEEEaftertitletext{%
  \noindent\begin{minipage}{\textwidth}
    \centering
    \FigureSlot{\textwidth}{8.0cm}{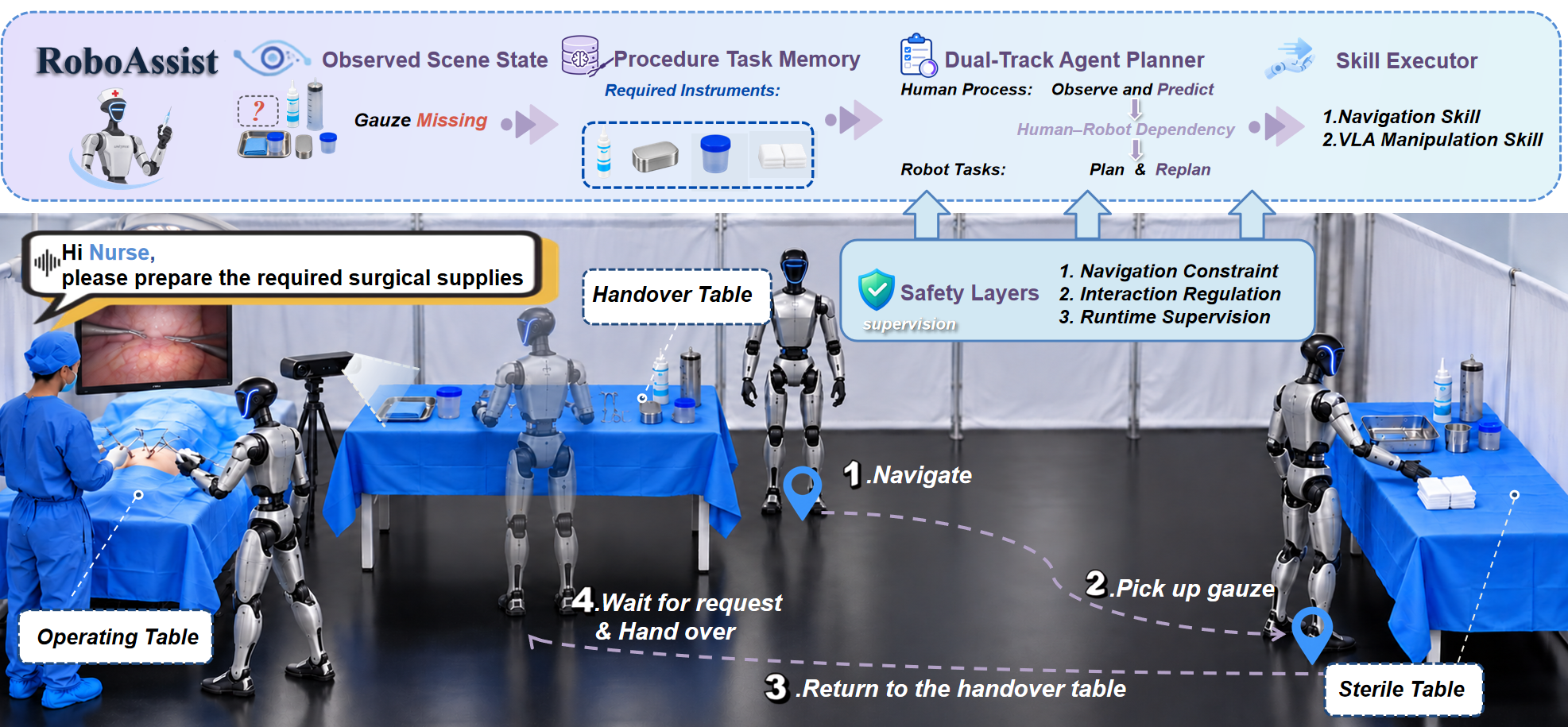}  
    \begin{CJK*}{UTF8}{gbsn}
    \captionof{figure}{RoboAssist perceives missing surgical items, retrieves task requirements from memory, plans and executes navigation and manipulation skills, and delivers requested supplies to the surgeon under continuous safety supervision.}
    \label{fig:arch}
    \end{CJK*}
  \end{minipage}
}

\begin{document}

\maketitle
\thispagestyle{empty}
\pagestyle{empty}
\raggedbottom 
\setlength{\textfloatsep}{6pt plus 1pt minus 2pt}

\begin{abstract}
Long-horizon surgical assistance requires humanoid robots to coordinate with evolving human activities while maintaining safety across planning and execution. We present RoboAssist, an agent-based framework for interactive human–humanoid planning that integrates workflow reasoning, task coordination, and cross-layer safety. At its core is an asymmetric dual-track representation that
separates partially observed human process states from executable robot
task sequences. By updating human-process estimates, scene context, and task
dependencies online, RoboAssist revalidates the remaining task
sequence and replans only the affected suffix when workflow
requests change. A cross-layer safety architecture combines preventive navigation
regulation, reactive regulation during close-range handover, and
independent whole-body runtime supervision. This design couples online task coordination with safety
constraints throughout execution. We demonstrate the framework on a Unitree G1 humanoid robot in long-horizon, multi-stage simulated surgical assistance scenarios encompassing multimodal interaction, instrument handling, medical material transport, navigation, and safe human–robot handover. Experiments show multi-stage task completion and adaptation to
workflow-request changes. A targeted full-replanning ablation shows that residual replanning
reduces plan-update latency and post-update token usage.
Separate safety experiments demonstrate complementary protection across
navigation, handover, and runtime supervision. Additional results and demonstrations are available online at
\url{https://roboassist.github.io}.
\end{abstract}


\section{INTRODUCTION}

Humanoid robots offer a promising platform for surgical assistance by
combining navigation and manipulation in human-designed environments.
Beyond isolated commands, an autonomous assistant must coordinate
instrument delivery, material transport, and handovers with an evolving
surgical workflow, requiring reasoning about human activity, assistance
needs, and timing. Over long horizons, workflow and scene changes can
invalidate task dependencies and prior plans; the central challenge is
therefore coordinated human--humanoid behavior with safety throughout
planning and execution~\cite{yang2017medical}.

Finite-state machines, behavior trees, and symbolic planners provide
explicit, modular execution but rely on predefined states, transitions,
predicates, and recovery logic~\cite{colledanchise2018behavior}.
In surgical assistance, manually specifying such logic becomes difficult
as human activities and task dependencies vary. Motion planning and
feedback control address physical feasibility, but effective assistance
also requires coupling geometric constraints with workflow context and
human activity.

Large language and multimodal models enable robots to interpret human
instructions and reason about complex tasks. Agent-based systems integrate
perception, memory, task decomposition, skill invocation, and feedback to
revise plans online. However, autonomous execution alone does not ensure
alignment with evolving human workflows or consistent safety across
planning and control.

First, planning must distinguish partially observed human process states
from executable robot tasks while continuously grounding their
dependencies in evolving interaction and scene context. Second, safety must remain consistent across planning, skill execution, close-range interaction, and whole-body control. Addressing both requires an explicit connection between interactive process reasoning and runtime safety constraints.

We present RoboAssist, an agent-based framework for long-horizon
human--humanoid surgical assistance, as illustrated in
Fig.~\ref{fig:arch}. RoboAssist separately maintains
partially observed human-process states and executable robot tasks,
linking them through evidence-gated dependencies. Observation-updated
task memory is used to revalidate these dependencies and trigger
residual replanning only for the affected task suffix. Task-level
decisions are translated into navigation and manipulation skills
through a unified execution interface, while safety constraints remain
active across planning, handover, and whole-body execution.

The main contributions of this work are as follows:

\begin{itemize}

\item We formulate long-horizon surgical assistance as interactive
human--humanoid planning under evolving workflow and scene states.
An asymmetric dual-track representation separates partially observed
human-process states from executable robot tasks and links them through
evidence gates, while observation-updated task memory enables
dependency-aware residual replanning of the affected task suffix.

\item We develop a cross-layer safety architecture that coordinates
surgeon-aware navigation regulation, reactive regulation during
close-range handover, and independent whole-body runtime supervision,
providing progressively stronger intervention across the autonomy
stack.

\item We implement RoboAssist on a Unitree G1 humanoid and evaluate it
in long-horizon, multi-stage simulated surgical assistance. Experiments
assess task execution, adaptation to workflow-request changes, planning
overhead through a targeted full-replanning ablation, and complementary
safety protection during navigation, manipulation, and human--robot
handover.

\end{itemize}


\section{Related Work}

\subsection{Humanoid-Robot Collaboration and Surgical Assistance}

Robotic surgical assistants increasingly automate instrument provisioning and exchange.
Quirubot combined speech recognition, visual localization, and pick-and-place for instrument delivery~\cite{perezvidal2012steps}, while Wagner \textit{et al.} used laparoscopic video to anticipate instrument needs~\cite{wagner2024anticipate}. Recent work has further explored surgical affordance prediction for robot autonomy~\cite{song2026surgam}.
Human--robot handover requires coordinated perception, planning, grasping, release, and interaction safety~\cite{ortenzi2021handovers}; Yang \textit{et al.} combined grasp reachability, contact detection, and model-predictive control for responsive handover~\cite{yang2022model}.
Humanoids have also entered surgical workspaces, including teleoperated Unitree G1 assistance for endoscopic visualization~\cite{cho2026humanoid}. Beyond surgical settings, recent work has demonstrated whole-body loco-manipulation and scene interaction on Unitree G1~\cite{yang2026omniretarget}.

RoboAssist instead addresses autonomous long-horizon coordination
across multi-stage surgical-assistance activities, grounding task
memory in workflow and scene observations to coordinate navigation,
manipulation, and human--robot handover.

\subsection{Foundation Models and Agent-Based Robotic Planning}

Foundation models increasingly bridge high-level instructions and executable robot behaviors~\cite{driess2023palme,wang2024hypermotion,liang2023code,wang2025intention}.
SayCan grounds language-model reasoning in learned affordances~\cite{ichter2022do},
Inner Monologue incorporates execution feedback for closed-loop planning~\cite{huang2022innermonologueembodiedreasoning},
and WildLMa extends skill-based planning to long-horizon loco-manipulation~\cite{qiu2025wildlma}.
SymSkill further combines symbolic planning with reusable skills for reactive long-horizon manipulation~\cite{shao2026symskill}.

Language-conditioned robot policies and VLA models map multimodal observations and instructions to robot actions.
RT-1 demonstrated large-scale language-conditioned robotic control~\cite{brohan2023rt1},
while RT-2 transfers vision-language knowledge to robotic control~\cite{zitkovich2023rt2}, OpenVLA provides an adaptable generalist policy~\cite{kim2025openvla}, and GR00T N1 extends foundation-model policies to humanoid manipulation~\cite{nvidia2025groot}.
RoboNurse-VLA further integrates voice interaction, visual perception, and VLA control for surgical instrument grasping and handover~\cite{li2025robonurse}.

RoboAssist complements these skill-based and foundation-model planners
by explicitly separating partially observed human-process states from
executable robot tasks, linking them through evidence-gated
dependencies, and updating only the affected task suffix when those
dependencies change.

\subsection{Humanoid Safety-Aware Planning and Control}

Safety in human--robot shared workspaces requires protection across
navigation, close-range interaction, and runtime execution~\cite{lasota2017safehri}.
Human-aware navigation incorporates human-centered spatial constraints~\cite{kruse2013humanaware}, while safe handover requires controlled robot and object motion near humans~\cite{ortenzi2021handovers}.
Proprioceptive sensing and collision monitoring further support abnormal interaction detection and safe responses~\cite{haddadin2017collisions}.

RoboAssist adopts a three-layer safety design: a surgeon-centered
protected operating region with proximity-based deceleration and
boundary-triggered stopping for navigation, reactive safety
regulation during VLA-based handover, and runtime supervision of
force and joint-motion anomalies, with detected violations
triggering an emergency stop.


\begin{figure*}[h]
    \centering
    \includegraphics[
        width=1.0\textwidth,
        trim={0cm 0cm 0cm 0cm}, 
        clip
    ]{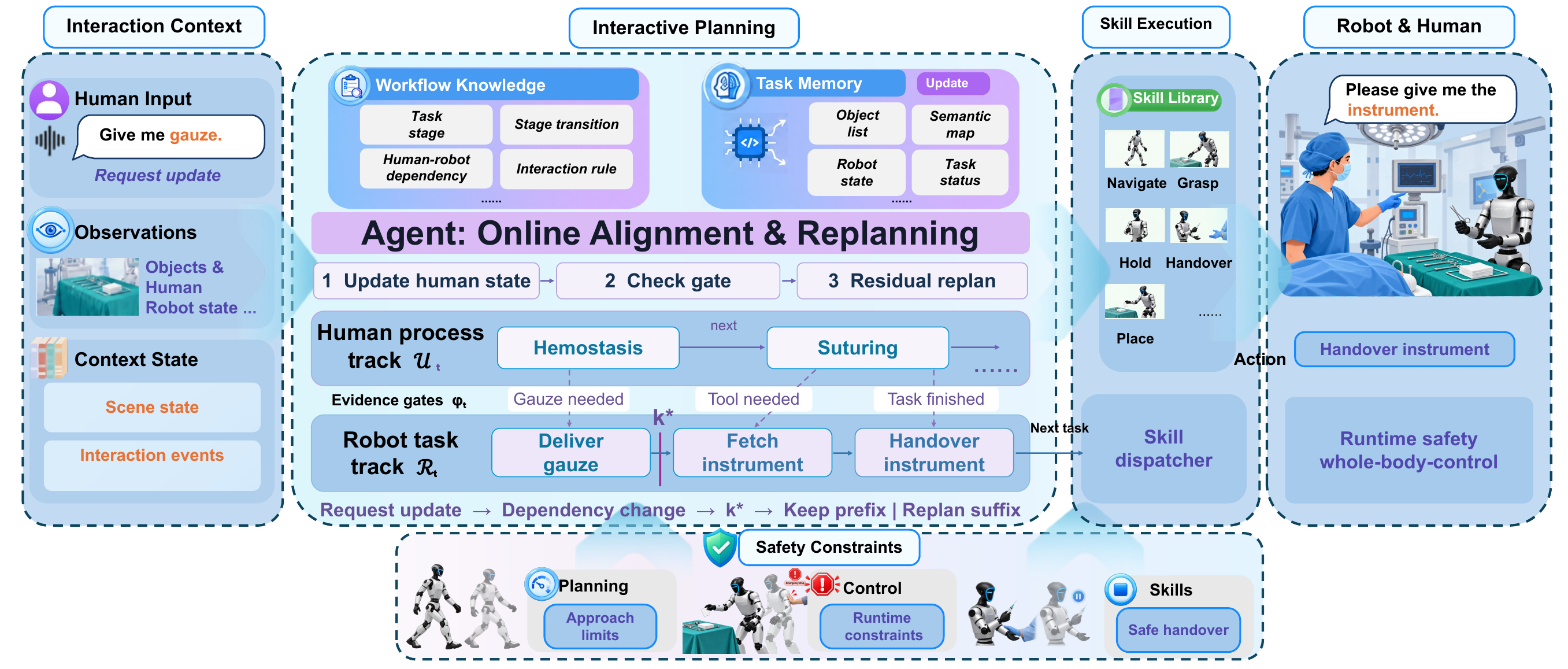}
    \caption{\textbf{The proposed framework integrates} workflow knowledge, task memory, scene observations, and human requests for online task alignment and replanning. The agent predicts human–robot task progression, dispatches executable skills, and enforces runtime safety constraints during navigation, manipulation, and handover.}
    
    \label{framework}
\end{figure*}

\section{Methodology}

\subsection{Task formulation and system overview}

\subsubsection{Long-Horizon Surgical Assistance}
We consider a humanoid assistant that prepares, transports, and
hands over task-relevant objects under evolving human instructions
and scene conditions. Later skills depend on earlier physical effects
and human readiness; therefore, changes in task requirements, object
locations, or interaction conditions may invalidate the remaining
plan. We focus on task-level planning, while low-level behaviors are
encapsulated as parameterized skills in the library $\mathcal{K}$.

\subsubsection{System Overview}
As shown in Fig.~\ref{framework}, RoboAssist comprises a
perception-and-memory module, an LLM-driven task-level Agent,
a skill executor, and a safety layer. Human and scene
observations update the estimated procedural state and task memory
$\mathcal{M}_t$. The Agent maintains the human process track
$\mathcal{U}_t$ and robot task track $\mathcal{R}_t$, while the executor
checks task preconditions and records execution outcomes.

At decision time $t$, the Agent operates on
\begin{equation}
\mathcal{C}_t =
\left\langle
s_t,\mathcal{Z}_t,\mathcal{K},\mathcal{M}_t,\mathcal{H}_t
\right\rangle .
\end{equation}
Here, $s_t$ denotes the current robot and scene state;
$\mathcal{Z}_t$ links human progress and robot tasks;
$\mathcal{K}$ is the skill library; $\mathcal{M}_t$ stores
observation-updated task and scene information; and $\mathcal{H}_t$
records instruction updates, skill invocations, and execution outcomes.

\subsection{Interactive planning and online process alignment}

RoboAssist separates partially observed human progress from executable
robot tasks and links them through explicit gating relations. This
avoids mixing non-controllable process estimates with robot-controllable
actions and supports dependency updates as the surgical workflow evolves.

\subsubsection{Asymmetric Dual-Track Representation}
Human progress is inferred from voice and visual observations and may
remain uncertain, whereas robot tasks are directly executable.
Accordingly, RoboAssist maintains the two tracks separately and
connects them through explicit gates without requiring strict temporal
synchronization.

\textbf{Human Process Track.}
We define $\mathcal{U}_t$ as a set of process nodes
representing observed or confirmed human states and
predicted next actions:
\begin{equation}
    u_i = (\eta_i,\ c_i,\ \tau_i).
\end{equation}
Here, $\eta_i$ describes the human state or action, associated entities,
and assistance need linked through $\mathcal{M}_t$; $c_i$ records whether
the estimate comes from voice or visual evidence or Agent inference; and
$\tau_i$ is its verifiable evidence criterion. The Agent estimates
procedural progress and predicts subsequent actions from observations,
workflow knowledge, and interaction history. Predictions may trigger task
generation or preparatory actions but cannot satisfy gates; a human-process
condition is confirmed only when $\tau_i$ is satisfied. Missing evidence
leaves it unconfirmed, whereas contradictory evidence invalidates prior
confirmation and dependent robot tasks.

\textbf{Robot Task Track.}
We define $\mathcal{R}_t$ as a sequence of robot task nodes:
\begin{equation}
    v_j = (k_j,\ \theta_j,\ \sigma_j,\ p_j,\ e_j)
\end{equation}
where $k_j \in \mathcal{K}$ is the bound skill, $\theta_j$ its
invocation parameters, $\sigma_j$ its execution status
(pending, executing, completed, failed, canceled), and $p_j$ and
$e_j$ denote its precondition and expected effect, respectively.
The Agent generates $\mathcal{R}_t$ from confirmed requirements and
anticipated assistance needs associated with observed and predicted
human-process states, using workflow knowledge, $\mathcal{M}_t$, and
skills in $\mathcal{K}$.

The task precondition $p_j(s)$ represents robot- and scene-dependent
physical conditions, such as object availability, object possession,
robot configuration, and resource availability. Human-process
readiness is evaluated separately through the gating relation below,
while safety constraints are enforced independently during execution.

\textbf{Gating Map.}
The two tracks are linked by a gating map:
\begin{equation}
    \mathcal{Z}_t = (\mathcal{U}_t,\mathcal{R}_t,\phi_t),
    \qquad
    \phi_t(j) \subseteq \{1,\dots,|\mathcal{U}_t|\},
\end{equation}
where $\phi_t(j)$ indexes the human-process nodes on which
task $v_j$ depends. One human-process node may gate multiple
robot tasks, such as ``disinfection completed'' enabling both
instrument and gauze delivery. An empty $\phi_t(j)$ indicates
that $v_j$ has no human-process gate. Such ungated tasks can
therefore be scheduled as preparatory actions before later
evidence-gated interaction tasks, provided that their physical
preconditions and safety constraints are satisfied.

The map $\phi_t$ provides traceable human--robot dependencies
for consistency checking and residual replanning.

\subsubsection{Cross-Track Consistency and Residual Replanning}

At each decision cycle, new voice and visual observations update
the human-process state and scene memory. Conditioned on these
updates, workflow knowledge, and interaction history, the Agent
revises $\mathcal{U}_{t+1}$ and checks the remaining robot tasks
for changes in requirements, dependencies, or scene conditions.
For each unfinished task $v_j$, gate satisfaction is evaluated as

\begin{equation}
    g_j(\mathcal{U}_{t+1})
    = \bigwedge_{i \in \phi_t(j)}
      \tau_i(\mathcal{U}_{t+1}),
    \label{eq:gating_condition}
\end{equation}

where $\tau_i(\mathcal{U}_{t+1})\in\{0,1\}$ indicates whether the
corresponding criterion is supported by admissible observational evidence
or explicit human confirmation. For $\phi_t(j)=\varnothing$, $g_j=1$.
An unsatisfied gate causes the task to wait, while physical preconditions
$p_j(s)$ are checked at dispatch. An unmet future precondition does not
invalidate the plan if a valid preceding task can establish it. Residual
replanning is triggered when new evidence invalidates a requirement or
human-process dependency, or when a required precondition can no longer
be established.

The divergence point $k^\ast$ is the earliest unfinished task whose
requirement or human-process dependency is invalidated, or whose
physical precondition can no longer be established by the retained
prefix. The prefix $\mathcal{R}_t^{<k^\ast}$ is retained, including
completed records and still-valid pending tasks, while mutable physical
facts such as object possession, robot location, and resource occupancy
are revalidated from the current state and memory before reuse.

If an executing task becomes invalid, the skill executor first
interrupts or cancels it and returns the resulting physical state.
The affected suffix is then regenerated using both the current
validated state and the retained pending prefix, so that the new
tasks remain compatible with the expected effects of the preserved
tasks. The corresponding
gates are updated, yielding
\begin{equation}
    \mathcal{Z}_{t+1}
    = \left(
        \mathcal{U}_{t+1},
        \mathcal{R}_{t+1},
        \phi_{t+1}
      \right).
    \label{eq:updated_dual_track}
\end{equation}

Algorithm~\ref{alg:interactive_planning} summarizes this
evidence-update, state-validation, and residual-replanning loop,
including execution subject to human-process gates, physical
preconditions, and safety constraints.

\begin{algorithm}[t]
\caption{Interactive Planning with Residual Replanning}
\label{alg:interactive_planning}
\small
\begin{algorithmic}[1]
\REQUIRE State $s$, memory $\mathcal{M}$, workflow knowledge,
skill library $\mathcal{K}$
\STATE Initialize $\mathcal{U},\mathcal{H},\mathcal{R}$, and $\phi$
\WHILE{the assistance session is active}
    \STATE Update $s,\mathcal{M},\mathcal{H},\mathcal{U}$, task statuses,
    and mutable facts
    \STATE Infer assistance needs and locate the earliest affected task $k^\ast$
    \IF{the executing task is invalidated}
        \STATE Safely interrupt it and update $s$
    \ENDIF
    \IF{$k^\ast \neq \varnothing$}
        \STATE Preserve $\mathcal{R}^{<k^\ast}$; regenerate the affected
        suffix; update $\phi$
    \ENDIF
    \IF{the executor is idle and runtime supervision permits motion}
        \STATE Let $v_j$ be the earliest pending task, or $\varnothing$
        \IF{$v_j \neq \varnothing$, $g_j(\mathcal{U})=1$, and $p_j(s)=1$}
            \STATE Execute $v_j$ under safety constraints
        \ENDIF
    \ENDIF
    \STATE Record feedback in $\mathcal{R},\mathcal{M},\mathcal{H}$
\ENDWHILE
\end{algorithmic}
\end{algorithm}

\subsection{Cross-Layer Safety Architecture}
\label{sec:safety_contract}

\begin{figure}[!t]
    \centering
    \includegraphics[
        width=\linewidth,
        keepaspectratio
    ]{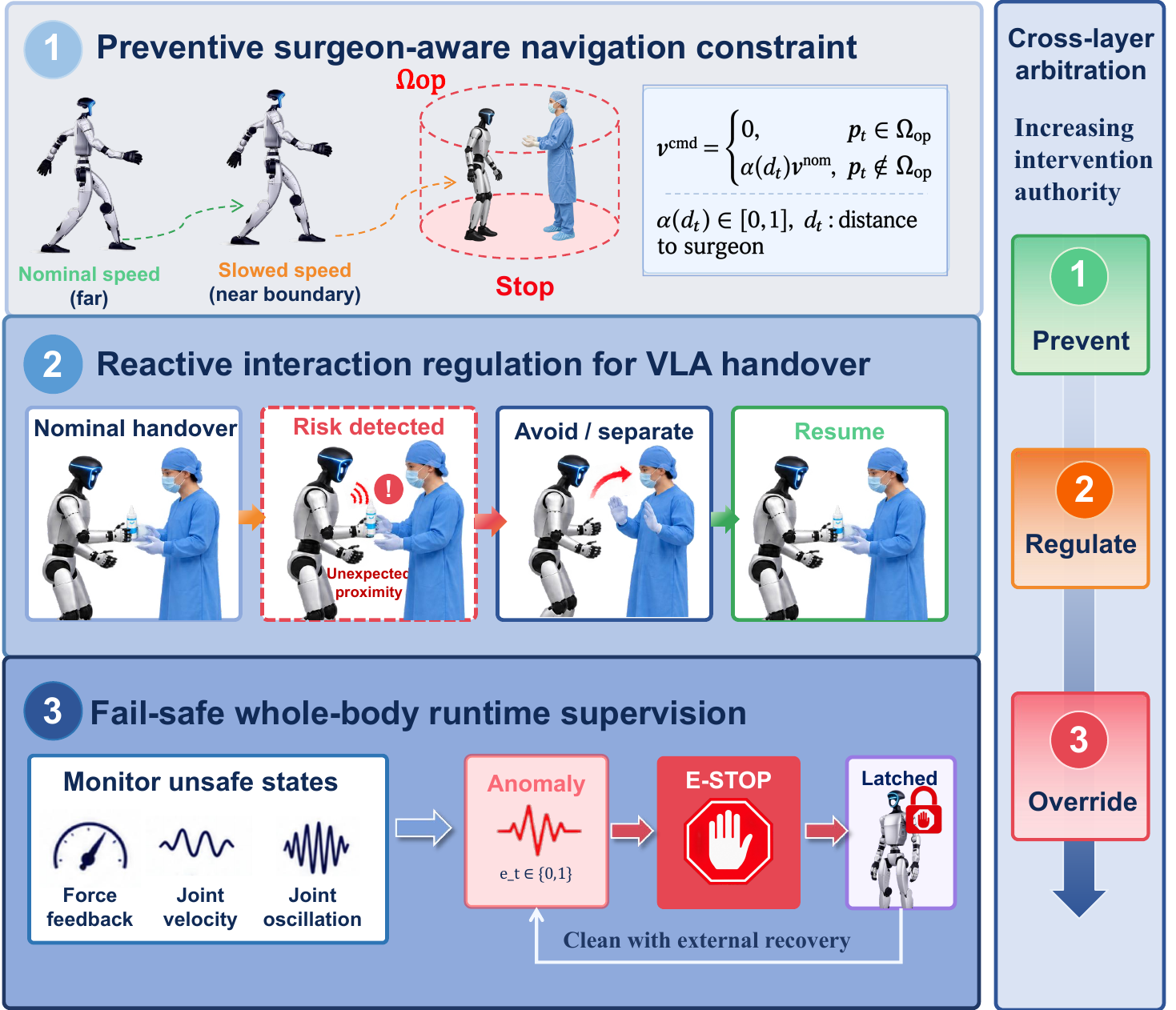}
    \caption{Cross-layer safety architecture combining surgeon-aware
navigation regulation, reactive handover regulation, and fail-safe
whole-body runtime supervision with increasing intervention authority.}
    \label{fig:safety_contract}
\end{figure}

As summarized in Fig.~\ref{fig:safety_contract}, RoboAssist uses
three safety layers with increasing intervention authority:
navigation regulation, interaction regulation, and fail-safe
runtime supervision.

\textbf{Preventive surgeon-aware navigation regulation.}
During locomotion, the commanded robot velocity is regulated
according to its clearance from the surgeon's protected operating
region. Let $\Omega_{\mathrm{op}}$ denote the protected region,
$v_t^{\mathrm{nom}}$ the nominal locomotion command, and $d_t$ the
signed clearance between the robot navigation reference and the
boundary of $\Omega_{\mathrm{op}}$, where $d_t>0$ denotes a position
outside the region and $d_t\leq 0$ denotes boundary contact or entry.
The executed command is

\begin{equation}
v_t^{\mathrm{cmd}} =
\begin{cases}
0, & d_t \leq 0,\\
\alpha(d_t)\,v_t^{\mathrm{nom}}, & d_t > 0,
\end{cases}
\label{eq:navigation_safety}
\end{equation}

where $\alpha(d_t)\in[0,1]$ is a clearance-dependent velocity
scaling factor. It approaches unity sufficiently far from the
protected region and decreases as the robot approaches its boundary.
Thus, the navigation layer progressively reduces the commanded
velocity near the protected region and issues a zero-velocity command
at boundary contact or entry.

The regulation is enforced at the skill-execution interface whenever
locomotion is enabled, including navigation and object transportation.
During close-range handover, manipulation motion is additionally
regulated by the interaction-safety layer described below.

\textbf{Reactive interaction regulation for VLA handover.}
Close-range object handover requires the robot manipulator to enter
the surgeon's reachable workspace. RoboAssist therefore separates
task-oriented motion generation from interaction safety regulation:
the VLA policy generates the nominal handover behavior, while a
VLM-based interaction monitor evaluates the relative human--robot
configuration during execution.

When unexpected proximity or human motion is detected as a potential
interaction risk, the nominal handover behavior is interrupted and an
avoidance response is issued to increase human--robot separation.
The nominal handover behavior can resume only after the interaction
condition is again classified as safe.

\textbf{Fail-safe whole-body runtime supervision.}
An independent runtime supervisor continuously monitors proprioceptive
feedback during navigation and manipulation. Abnormal force feedback,
excessive joint velocity, and persistent joint oscillation are treated
as unsafe execution states.

The anomaly flag combines the deployed force, joint-velocity, and
joint-oscillation detectors:
\begin{equation}
e_t = e_t^{F} \lor e_t^{V} \lor e_t^{O},
\label{eq:anomaly_flag}
\end{equation}
where $e_t^{F}$, $e_t^{V}$, and $e_t^{O}$ denote the binary outputs
of the force, joint-velocity, and persistent-oscillation detectors,
respectively. Each detector is evaluated using its fixed runtime
threshold and, where applicable, a persistence criterion.

Let $\ell_t\in\{0,1\}$ denote the latched emergency-stop state
and $r_t^{\mathrm{ext}}\in\{0,1\}$ an explicit external recovery
signal. The emergency state evolves as
\begin{equation}
\ell_t =
e_t \lor
\left(
\ell_{t-1} \land \neg r_t^{\mathrm{ext}}
\right),
\label{eq:emergency_latch}
\end{equation}
where $\lor$, $\land$, and $\neg$ denote logical OR, AND, and NOT,
respectively. Once activated, the supervisor terminates the current
skill, inhibits subsequent motion commands, and keeps the robot stopped
until explicit external recovery is issued.

\textbf{Safety arbitration.}
Navigation regulation modifies locomotion commands, while interaction
regulation may override nominal handover motion. The runtime
supervisor has the highest authority: its latched stop overrides
both layers and prevents further task-motion execution until
explicit external recovery.

\section{Experiment and Evaluation}
\label{sec:experiments}

\subsection{Experiment setup}
\begin{figure}
    \centering
    \includegraphics[width=\columnwidth]{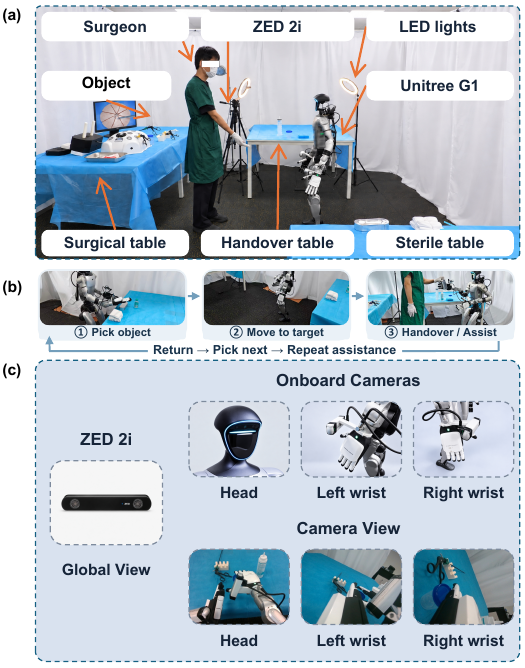}
    \caption{Experimental setup.
    (a) Overall experimental environment, including the surgeon, Unitree G1, surgical, handover, and sterile tables, global ZED 2i camera, and LED lighting. (b) Three task stages: picking, handover, and surgical assistance. (c) Camera configuration and representative views from the global camera and the robot’s head and wrist cameras.}
    \label{fig:experiment_setup}
\end{figure}

As shown in Fig.~\ref{fig:experiment_setup}, the workspace contains
an operating table, a handover table, and a sterile table for surgery,
object transfer, and material storage, respectively. The robot retrieves required objects from the sterile
table, transfers them via the handover table, and assists the surgeon
at the operating table.

The platform is a Unitree G1 humanoid with Inspire FTP dexterous hands.
A global ZED 2i and onboard head and wrist cameras provide
640$\times$480 video at 30 fps under two fill lights; onboard computation
uses an NVIDIA Jetson Thor. Spoken interaction, visual inventory perception, and task-level
decision-making and replanning use Qwen3.5-Omni-Plus-Realtime,
Qwen3.7-Plus, and DeepSeek V4 Pro, respectively.
Manipulation uses a shared GR00T N1.7 backbone with object-specific
grasp Action Experts and shared placement and handover experts.
The evaluated object set includes bottles, blue-capped cups, surgical
instruments, syringes, gauze, surgical drapes, and aluminum boxes,
spanning tools, consumables, containers, and transport items.

\subsection{Autonomous agent-based interactive planning}
To evaluate task execution and adaptation, we compare two baselines
and one planning ablation. \textbf{Scripted FSM} uses manually defined
states, transitions, retries, and safety-stop branches.
\textbf{Reactive agent} shares RoboAssist's observations, memory,
workflow knowledge, skills, and safety mechanisms, but uses a flat
task context and regenerates the remaining plan after relevant changes.
\textbf{RoboAssist} instead uses the asymmetric dual-track
representation, evidence gates, and $k^\ast$-based residual replanning.

\textbf{FullReplan}, our ablation variant, retains RoboAssist's
representation, gates, memory, and execution history but regenerates
all unfinished tasks after a request update. It does not repeat
completed actions. Agent model, perception, skills, safety mechanisms,
retry limits, and task budget are fixed across the three LLM-based
methods. Thus, the Reactive-agent comparison evaluates the combined effect
of the dual-track representation and evidence gating under
remaining-plan regeneration, whereas FullReplan versus RoboAssist
isolates residual suffix replanning.

\subsubsection{Long-horizon surgical assistance}
\begin{figure}[!b]
    \centering
    \includegraphics[width=0.8\columnwidth]{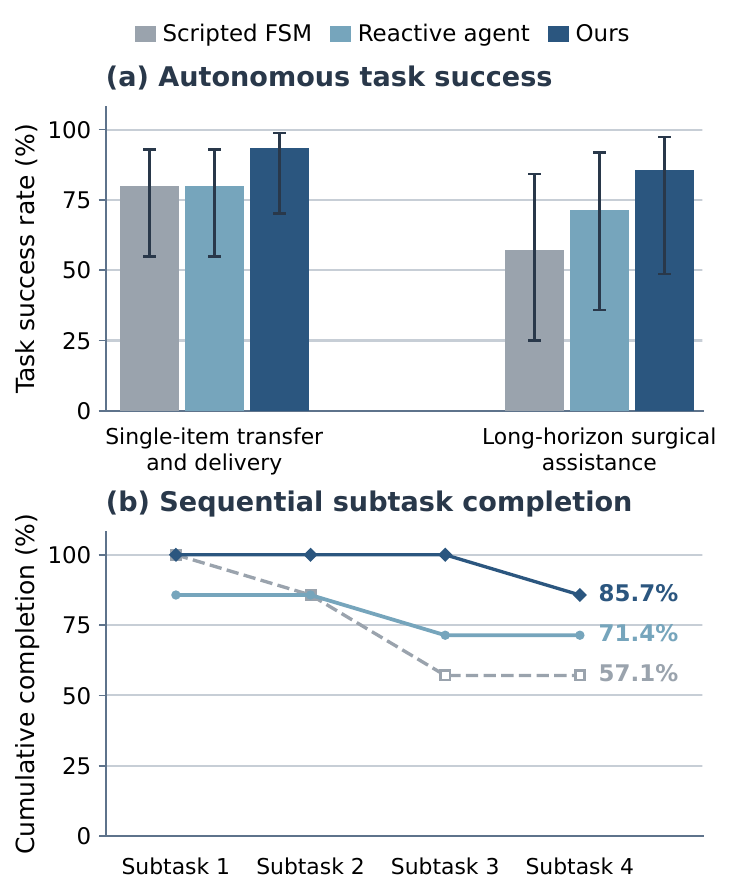}
    \caption{Task execution performance.
    (a) Success rates over 15 single-object trials and seven multi-object
    task trials per method, with 95\% Wilson confidence intervals.
    (b) Cumulative completion at four ordered subtask checkpoints in
    the multi-object task. Each point is the fraction of all seven
    trials that successfully complete every checkpoint up to and
    including the indicated subtask. The final point equals the
    corresponding multi-object task success rate in (a).}
    \label{comparison}
\end{figure}

We first assess single-object task reliability for the scripted
FSM, reactive agent, and RoboAssist across diverse objects in
15 trials per method. Success is defined as handing over the correct
object, and overall success rates are reported in
Fig.~\ref{comparison}(a).

We then evaluate long-horizon multi-object tasks in seven trials
per method. Each task combines navigation, alignment, grasping,
transportation, and handover, with later actions depending on the
states and progress established by preceding actions. Task success
requires completing all requested deliveries in the required order.

\begin{figure*}[h]
    \centering
    \includegraphics[width=\textwidth]{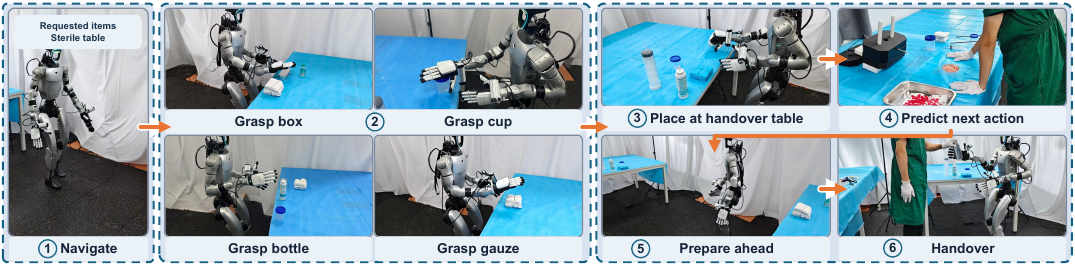}
    \caption{Representative moments of the RoboAssist
workflow: anticipating the surgeon's next action, navigating
to the object table, grasping and holding an object, handing
it over to the surgeon, and placing an object on the table.}
    \label{fig:workflow}
\end{figure*}

Fig.~\ref{comparison}(b) reports cumulative completion at four
ordered subtask checkpoints (Subtasks~1--4) within the multi-object
task. The value at Subtask~$j$ ($j=1,\ldots,4$) is the fraction of
all seven trials that successfully complete every checkpoint up to
and including that subtask. The final checkpoint marks task
completion, so its value equals the multi-object task success rate
in Fig.~\ref{comparison}(a). Representative execution stages are
shown in Fig.~\ref{fig:workflow}.

\subsubsection{Human-humanoid coordination and adaptation}

We evaluate execution efficiency in an independent multi-object
experiment using five successful runs per method, as shown in
Fig.~\ref{dynamic_task}.

For request-order adaptation, the scripted FSM, reactive agent,
FullReplan, and RoboAssist are each evaluated in 16 matched trials.
Each trial contains one request update that changes the required
delivery order, and succeeds only if all deliveries are completed
accordingly within the task budget without unplanned assistance.

Replanning latency is measured from planner acceptance of the request
update to commitment of the revised task queue, excluding speech
recognition and physical execution. Post-update token usage sums
task-planning input and output tokens from request acceptance to trial
termination, excluding initial planning and perception calls.
Table~\ref{tab:planning_ablation} reports success rates and planning
overhead; latency and token statistics use 10 logged trials per
LLM-based method.

\noindent
\begin{minipage}{\columnwidth}
    \centering

    \captionof{table}{Request-order adaptation and planning overhead.
    Success rates use 16 trials per method; latency and tokens are
    mean $\pm$ sample SD over 10 logged trials per LLM-based method.
    $\mathrm{k}=10^3$ tokens; N/A denotes not applicable.}
    \label{tab:planning_ablation}

    \begingroup
    \setlength{\tabcolsep}{2pt}
    \renewcommand{\arraystretch}{1.10}
    \begin{tabular*}{\columnwidth}{@{\extracolsep{\fill}}lccc@{}}
        \toprule
        \textbf{Method} &
        \makecell{\textbf{Success}\\(\%)$\uparrow$} &
        \makecell{\textbf{Replanning}\\\textbf{latency (s)}$\downarrow$} &
        \makecell{\textbf{Post-update}\\\textbf{tokens (k)}$\downarrow$} \\
        \midrule
        Scripted FSM & 0.00 & N/A & N/A \\
        Reactive agent & 56.25 &
        $3.97 \pm 0.52$ & $20.77 \pm 5.68$ \\
        FullReplan (ours) & 68.75 &
        $3.62 \pm 0.70$ & $20.02 \pm 3.84$ \\
        \textbf{Ours} & \textbf{81.25} &
        $\mathbf{1.20 \pm 0.21}$ & $\mathbf{11.41 \pm 2.98}$ \\
        \bottomrule
    \end{tabular*}
    \endgroup

    \vspace{0.3em}

    \includegraphics[width=\columnwidth]{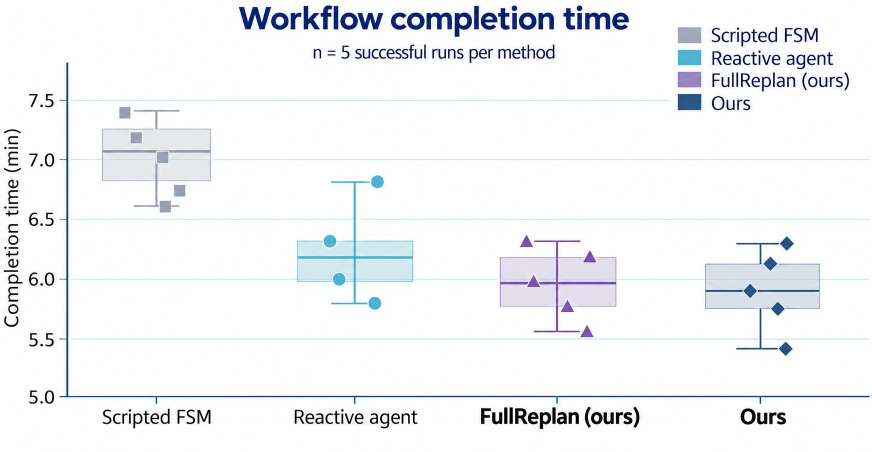}

    \vspace{-0.35em}

    \captionof{figure}{Completion times for successful multi-object workflows,
    with five successful runs per method. Markers denote individual runs.}
    \label{dynamic_task}

\end{minipage}
\par

\subsubsection{Cross-layer safety evaluation}

We evaluate navigation regulation through deceleration and stopping
near the protected region. The interaction and runtime layers are
evaluated in five close-range and five visual-blind-zone trials.
Response time is defined as the elapsed time from the hazard trigger
to the corresponding protective response and is calculated from
recorded timestamps for the VLM and force-feedback pathways.
An L3-OFF ablation evaluates the contribution of runtime supervision
under visual-blind-zone conditions.

\begin{figure}[t]
    \centering
    \includegraphics[width=\columnwidth]{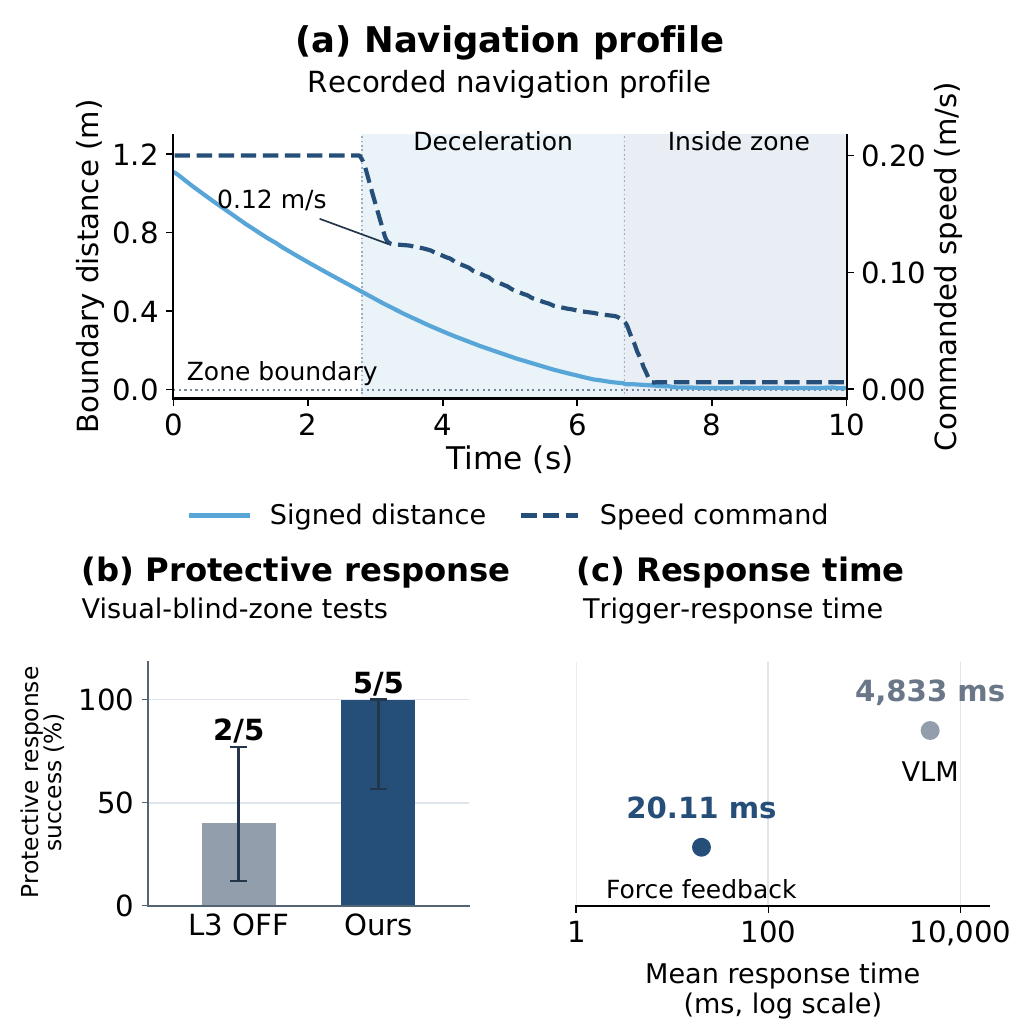}
    \caption{Cross-layer safety evaluation.
(a) Recorded navigation profile, where negative distance indicates
boundary crossing.
(b) Protective responses with 95\% Wilson confidence intervals;
L3 OFF disables the runtime supervisor.
(c) Mean response times of the VLM and force-feedback pathways,
measured from the corresponding hazard trigger to the protective response.}
    \label{embodiment}
\end{figure}

\subsection{Results and failure analysis}

\textbf{Task execution and efficiency.}
As shown in Fig.~\ref{comparison}(a), RoboAssist achieves a 93.3\%
single-object success rate, compared with 80.0\% for both the scripted
FSM and reactive agent. In the multi-object task evaluation, the
scripted FSM, reactive agent, and RoboAssist achieve 57.1\%, 71.4\%,
and 85.7\% success, respectively. Fig.~\ref{comparison}(b) shows
that RoboAssist maintains 100\% cumulative completion through the
first three subtask checkpoints and 85.7\% at the final checkpoint.
Fig.~\ref{dynamic_task} separately reports completion times from
five successful runs per method in the independent multi-object
task efficiency experiment.

\textbf{Request-order adaptation and ablation.}
As shown in Table~\ref{tab:planning_ablation}, the scripted FSM,
reactive agent, FullReplan, and RoboAssist achieve 0\%, 56.25\%,
68.75\%, and 81.25\% success, respectively. These rates measure end-to-end workflow
completion, including physical execution, rather than planning
correctness alone. Relative to FullReplan, RoboAssist reduces mean
replanning latency from 3.62 to 1.20~s (66.9\%).

Under the specified token accounting, mean post-update usage decreases
from 20.02k to 11.41k tokens (43.0\%).
These results support lower planning overhead from residual replanning;
the reactive-agent comparison reflects the combined planning
formulation rather than any single component.

\textbf{Cross-layer safety.}
As shown in Fig.~\ref{embodiment}(b), enabling the runtime supervisor
increases protective-response success from 40\% to 100\% in the
visual-blind-zone tests, supporting its complementary role when
visual protection is insufficient. Fig.~\ref{embodiment}(c) reports pathway-specific
trigger-to-response times of 20.11~ms for force feedback and
4.833~s for the VLM pathway. Because the two pathways use their
respective hazard triggers, these values characterize pathway
response times rather than a strictly matched end-to-end latency
comparison. The millisecond-scale force response therefore provides
a fast complementary protection layer during execution.

\textbf{Failure analysis.}
RoboAssist maintains 100\% cumulative completion through the first
three subtask checkpoints but decreases to 85.7\% at the final
checkpoint. This localizes the observed performance drop to the
last stage of the evaluated workflow, but does not by itself identify
whether the underlying cause is planning or physical execution.
Improving robustness at the final subtask therefore remains important
for long-horizon assistance.

\section{Conclusion}

We presented RoboAssist, an interactive planning framework
for long-horizon human--humanoid surgical assistance.
Its asymmetric dual-track representation connects human
process estimates with executable robot tasks, enabling
online dependency alignment and residual replanning.
A cross-layer safety architecture combines navigation regulation,
interaction protection, and independent runtime supervision. Experiments on a Unitree G1 in a simulated surgical-assistance setting
demonstrate multi-stage task execution and adaptation to workflow
requests. In the request-order-change benchmark, RoboAssist achieves a higher
observed success rate than FullReplan while reducing plan-update
latency and post-update token usage. Separate safety tests
demonstrate complementary protection under visual blind zones. Future work will extend the evaluation to longer and more
diverse workflows, broaden interaction and environmental
variations, and improve VLA skill robustness through
broader training coverage.



\bibliographystyle{IEEEtran}
\bibliography{references}








\end{document}